\documentclass{article}
\usepackage[hyphens]{url}
\usepackage{textcomp}
\usepackage{pifont}
\usepackage{array}
\usepackage{overpic}
\usepackage{tikz}
\usepackage[outline]{contour}
\usepackage{tcolorbox}
\usepackage{subfigure}
\usepackage{verbatim}
\usepackage{soul}
\usepackage{spconf,amsmath,graphicx,amssymb}
\usepackage{todonotes}
\usepackage{makecell}
\usepackage{color}
\usepackage{enumitem}
\usepackage{booktabs}
\usepackage{hyperref}
\usepackage[utf8]{inputenc}
\usepackage{lipsum}

\newcommand{\note}[1]{\textcolor{blue}{#1}}
\newcommand{\noter}[1]{\textcolor{red}{#1}}

\newcommand\copyrighttext{%
  \footnotesize \textcopyright 2024 IEEE. Published in 2024 IEEE International Conference on Image Processing (ICIP), scheduled for 27-30 October 2024 in Abu Dhabi, United Arab Emirates. Personal use of this material is permitted. However, permission to reprint/republish this material for advertising or promotional purposes or for creating new collective works for resale or redistribution to servers or lists, or to reuse any copyrighted component of this work in other works, must be obtained from the IEEE. Contact: Manager, Copyrights and Permissions / IEEE Service Center / 445 Hoes Lane / P.O. Box 1331 / Piscataway, NJ 08855-1331, USA. Telephone: + Intl. 908-562-3966.}
    \newcommand\mycopyrightnotice{%
\begin{tikzpicture}[remember picture,overlay]
\node[anchor=south,yshift=10pt] at (current page.south) {\fbox{\parbox{\dimexpr\textwidth-\fboxsep-\fboxrule\relax}{\copyrighttext}}};
\end{tikzpicture}%
}

\title{Improving Image de-raining using Reference-guided Transformers}
\name{Zihao Ye$^{1}$, Jaehoon Cho$^{2}$, Changjae Oh$^{1}$ }
\address{$^{1}$School of Electronic Engineering and Computer Science, Queen Mary University of London, UK \\
$^{2}$Hyundai Motor Company, Seoul, Korea }

\begin{document}
% \ninept
%
\maketitle
\begin{abstract}
% \JH{DRAFT(JH)}

Image de-raining is a critical task in computer vision to improve visibility and enhance the robustness of outdoor vision systems. While recent advances in de-raining methods have achieved remarkable performance, the challenge remains to produce high-quality and visually pleasing de-rained results. In this paper, we present a reference-guided de-raining filter, a transformer network that enhances de-raining results using a reference clean image as guidance. We leverage the capabilities of the proposed module to further refine the images de-rained by existing methods. We validate our method on three datasets and show that our module can improve the performance of existing prior-based, CNN-based, and transformer-based approaches.

% The core of our method lies in the integration of the Learning pattern Transformer Network into the de-raining pipeline. The LTTN, known for its ability to capture intricate pattern details and enhance visual appeal, is utilized to refine the de-rained images by learning and transforming their pattern patterns. By applying the LTTN's pattern transformation, we aim to enhance the perceptual quality of de-rained images, resulting in more visually accurate representations of rain-affected scenes.

% To evaluate the effectiveness of our approach, we conducted extensive experiments on diverse de-raining benchmarks. We not only assessed the visual quality improvements achieved by our method but also investigated its impact on downstream recognition tasks. The results demonstrate the potential of our method to not only enhance de-rained images but also to positively influence subsequent recognition accuracy.

% In summary, this paper presents a novel strategy for enhancing de-rained images through the integration of the Learning pattern Transformer Network. By refining the de-rained results, we aspire to not only achieve visually improved images but also boost the performance of recognition tasks relying on these images.
% \JH{DRAFT(JH)}
\end{abstract}
\mycopyrightnotice
\section{Introduction}
\label{sec:intro}
% \COcomment{As this is a short 4-page paper, we can include the related works in the introduction section.}

Image de-raining is an essential task in computer vision as rain streaks can decrease visibility and deteriorate the robustness of most outdoor vision systems. De-raining has been widely applied in a wide range of practical applications, including autonomous driving~\cite{huang2021memory,guo2021efficientderain} and surveillance systems~\cite{li2021online,li2021comprehensive}, as an essential pre-processing step. 

% Image de-raining can be divided into video-based and single image-based methods.
% Unlike video-based methods that can leverage the relationship between the image sequence, image-based methods are more challenging as no prior is given for the enhancement.

% Image de-raining is a critical task in the domain of computer vision, specifically within low-level vision tasks, aimed at restoring images degraded by rain streaks. This task holds significant potential for a wide array of practical applications, including but not limited to autonomous driving and pedestrian detection. The objective of single image de-raining is to construct a robust framework capable of performing the de-raining transformation using only a single rainy image as input.

Early approaches that solve the task with hand-crafted priors such as sparse coding~\cite{kang2011automatic} and Gaussian mixture model~\cite{li2016rain} are formulated to explicitly model the physical characteristics of rain streaks. However, they often fail under complex rain conditions and show over-smoothed images~\cite{cho2022memory}.
% With the advent of the Convolutional Neural Networks (CNNs),  single image de-raining has been investigated with CNNs, gaining significant improvements~\cite{Li_2020_CVPR,WANG2021106595,8627954,8767931,ren2019progressive}.
% By nature of CNNs, as the size of the network receptive field is limited, the pixel value inference of each spatial location only relies on small local surrounding regions~\cite{liang2021swinir}. Moreover, due to the ignorance of long-distance spatial context modeling, CNN-based methods often have difficulty accurately detecting heavy rain streaks, resulting in an often overly blurred result~\cite{li2018non}.
% To alleviate such limitations, more recently, transformer-based methods have emerged as a promising alternative ~\cite{9577359,Liang_2022_CVPR,li2021localvit,Chen_2023_CVPR,Wang_2022_CVPR}.
% While they can better model the non-local information for high-quality image reconstruction, the image details, which are local features of images, are not modeled well by these approaches when restoring clear images~\cite{Chen_2023_CVPR}. This stems from the self-attention in Transformers that does not model the local invariant properties, unlike CNNs. Therefore, while the single image de-raining methods have significant achievements, they still have much room for improvement.
The advent of Convolutional Neural Networks (CNNs) has led to substantial advances in single image de-raining~\cite{Li_2020_CVPR, WANG2021106595,8627954,8767931,ren2019progressive,cho2020single}. CNNs, however, have limited receptive fields, which means that the pixel value estimation for each spatial location primarily depends on small local surroundings. Therefore, due to the limited capacity for modeling long-range spatial context~\cite{liang2021swinir}, CNN-based methods often struggle with accurately detecting heavy rain streaks, leading to blurred results~\cite{li2018non}.
Transformer-based methods~\cite{9577359,Liang_2022_CVPR,li2021localvit,Chen_2023_CVPR,Wang_2022_CVPR} have emerged as a promising alternative as they can better capture non-local information, enhancing image reconstruction quality. However, these approaches do not model local image details well, which are crucial for achieving clear image restoration~\cite{Chen_2023_CVPR}. This limitation arises from the self-attention mechanism in Transformers, which does not adequately handle the local invariant properties, in contrast to CNNs.
While single image de-raining methods have made significant progress, there remains room for improvement in their ability to handle diverse and challenging rain conditions.\footnote{More results and the code: \url{http://ziiihooo.com/blog/2024/derain/}}

% In this paper, we introduce a reference-guided de-raining filter designed. Unlike traditional approaches that directly transform rainy images to clean ones, our model specializes in the transfer of useful patterns. Given a baseline de-raining model and an image retrieval method, we initially retrieve a reference image. Subsequently, we obtain de-rained versions of both the original and the reference images using the baseline model. Our pipeline aims to identify similar patterns between these de-rained images, as their correlation is more representative due to the shared experience of rain streak degradation and subsequent recovery by the baseline model. Once useful patterns are identified, they are transferred from the reference clean image, which is devoid of rain streak degradation and thus contains richer patterns. This nuanced methodology has shown to enhance the performance of existing de-raining models across a variety of datasets, especially those captured in environments with similar scenes~\footnote{ More results are available at: \url{http://ziiihooo.com/blog/2024/derain/}}.

\begin{figure}[!t]
  \centering
  \renewcommand{\arraystretch}{0.5}
  \setlength{\tabcolsep}{2pt} 
  \begin{tabular}{c@{\hspace{1pt}}c}
    \includegraphics[width=0.45\columnwidth,height=0.06\textheight]{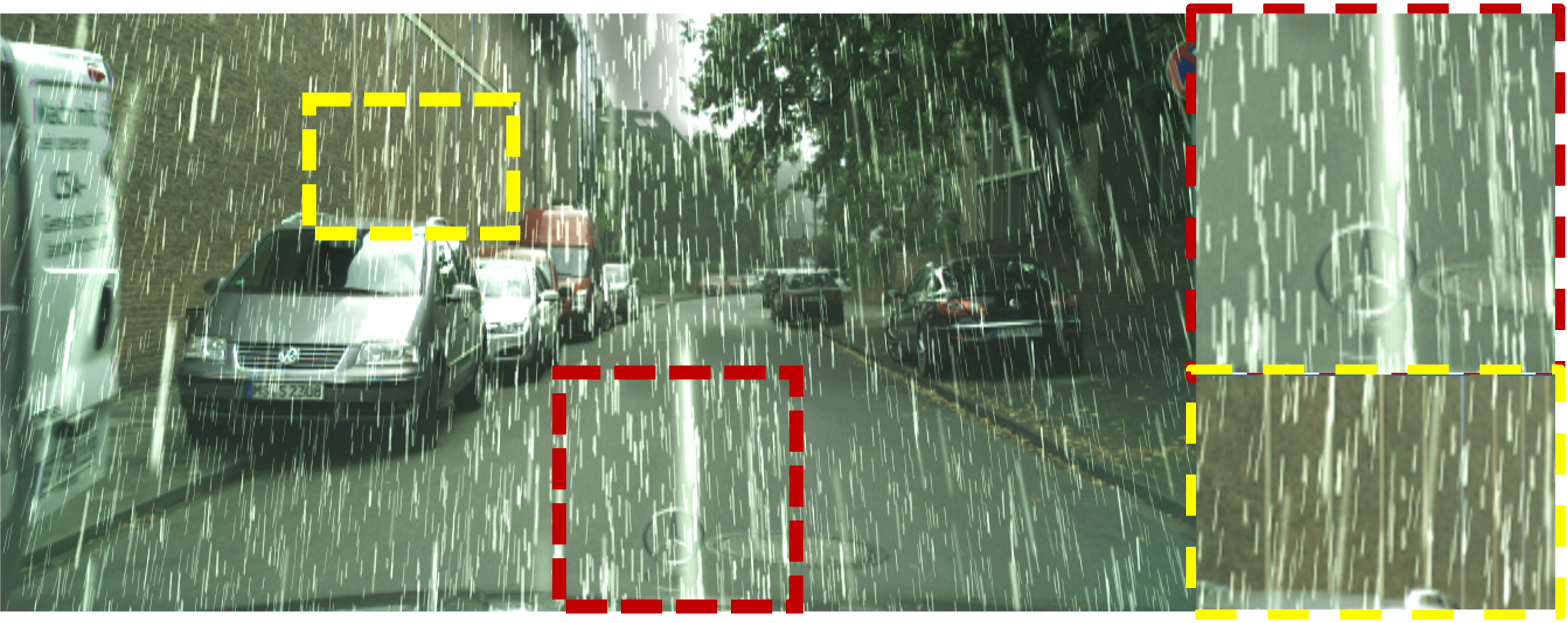} &
    \includegraphics[width=0.45\columnwidth,height=0.06\textheight]{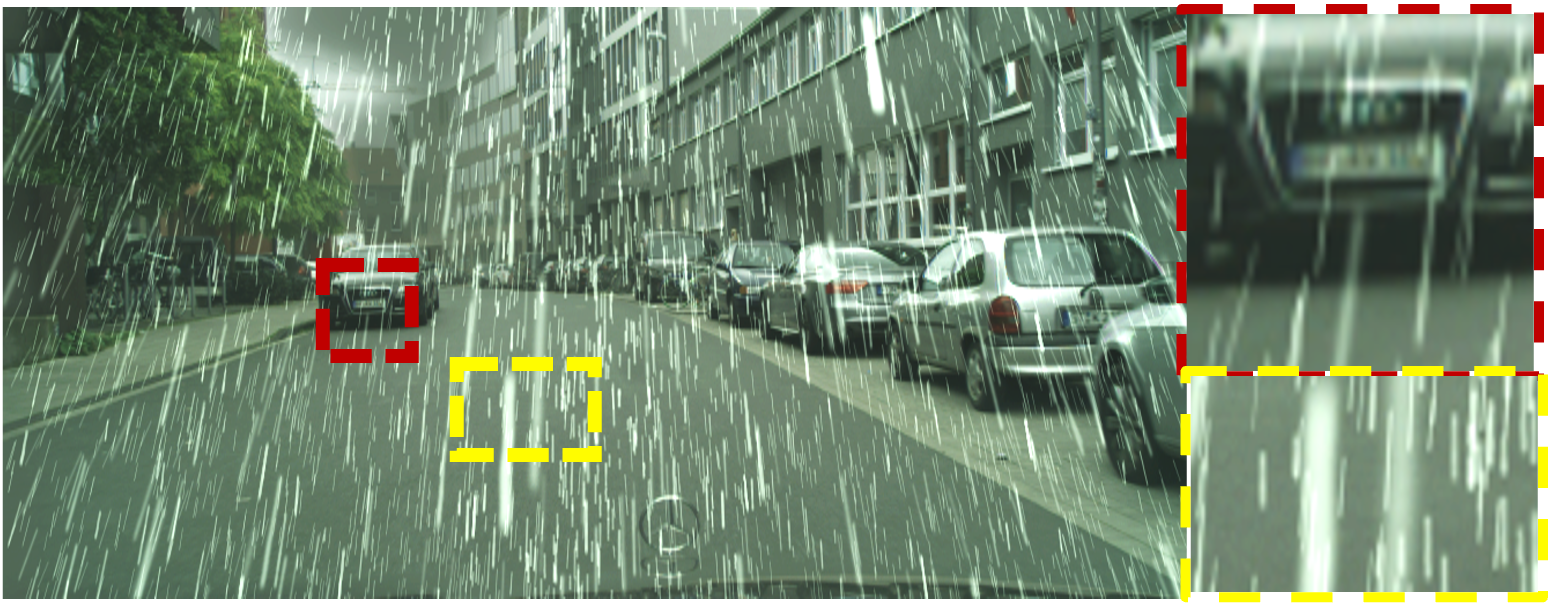}  \\
    \multicolumn{2}{c}{Rainy images}\\
    \includegraphics[width=0.45\columnwidth,height=0.06\textheight]{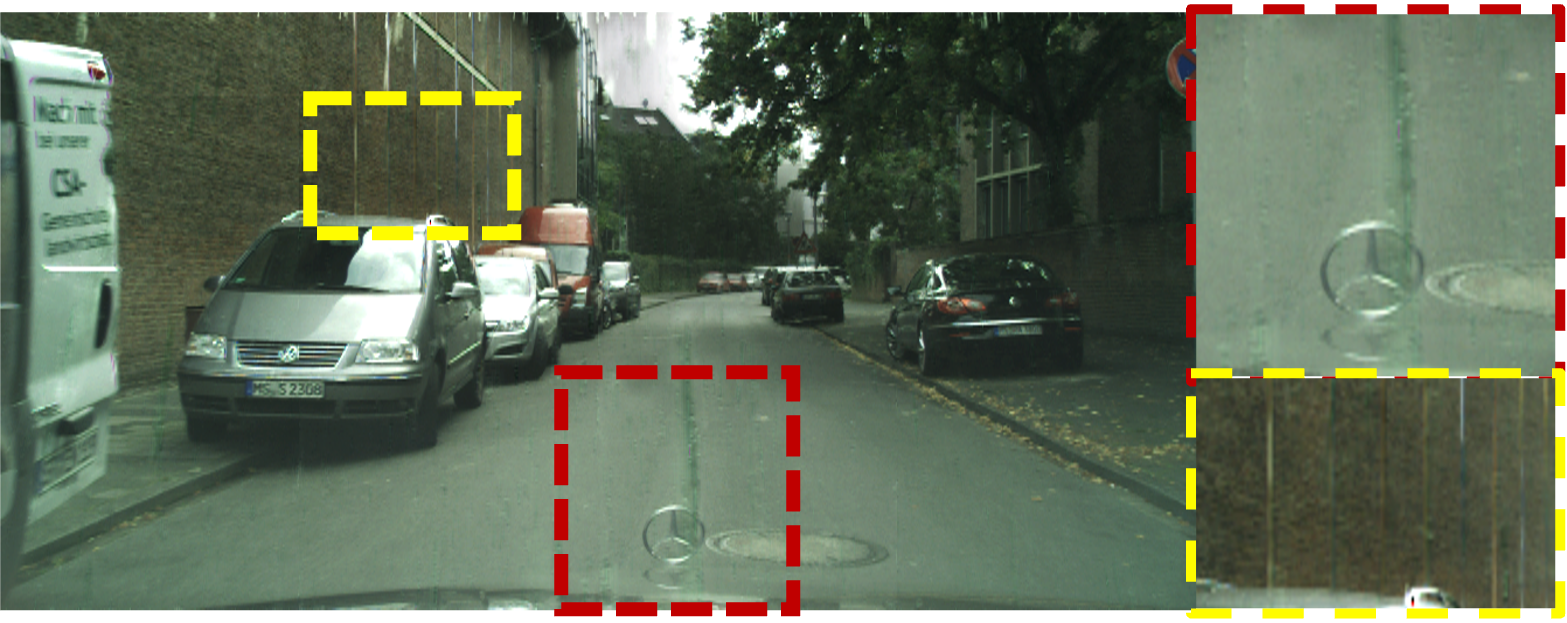} &
    \includegraphics[width=0.45\columnwidth,height=0.06\textheight]{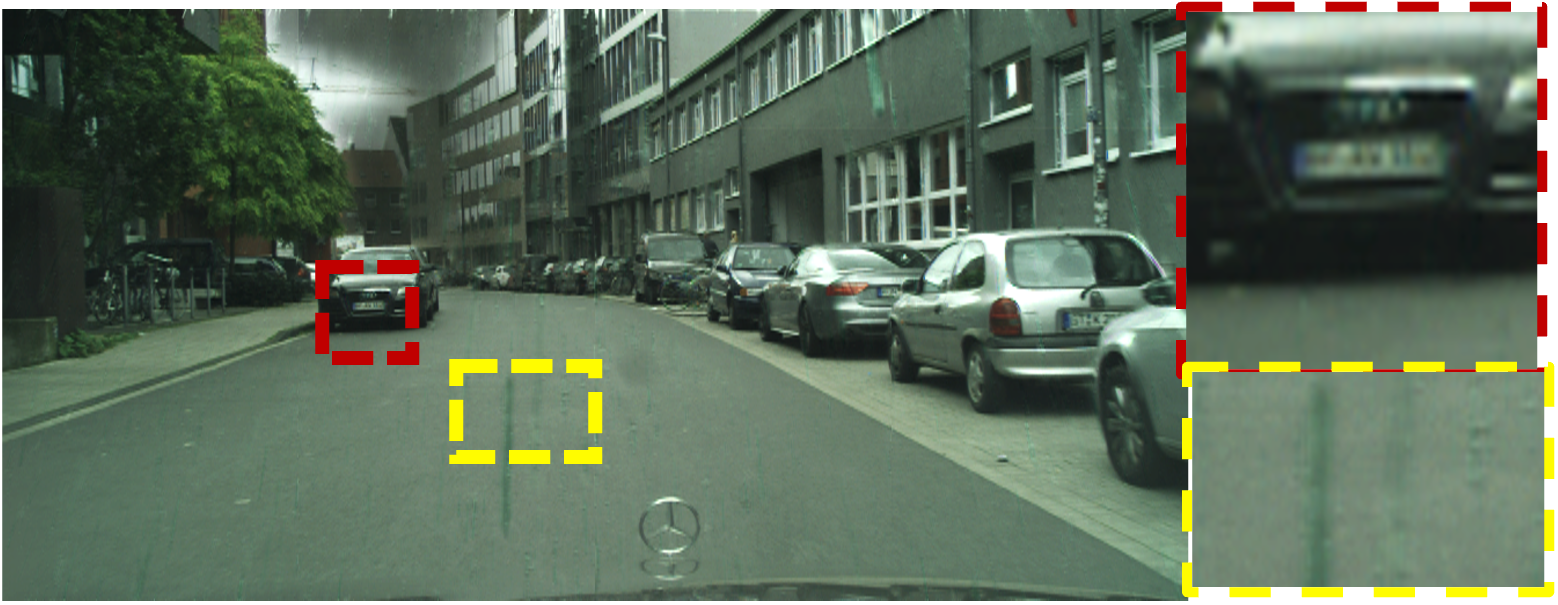}\\
    \multicolumn{2}{c}{De-rained results obtained by Uformer~\cite{Wang_2022_CVPR}}\\
    \includegraphics[width=0.45\columnwidth,height=0.06\textheight]{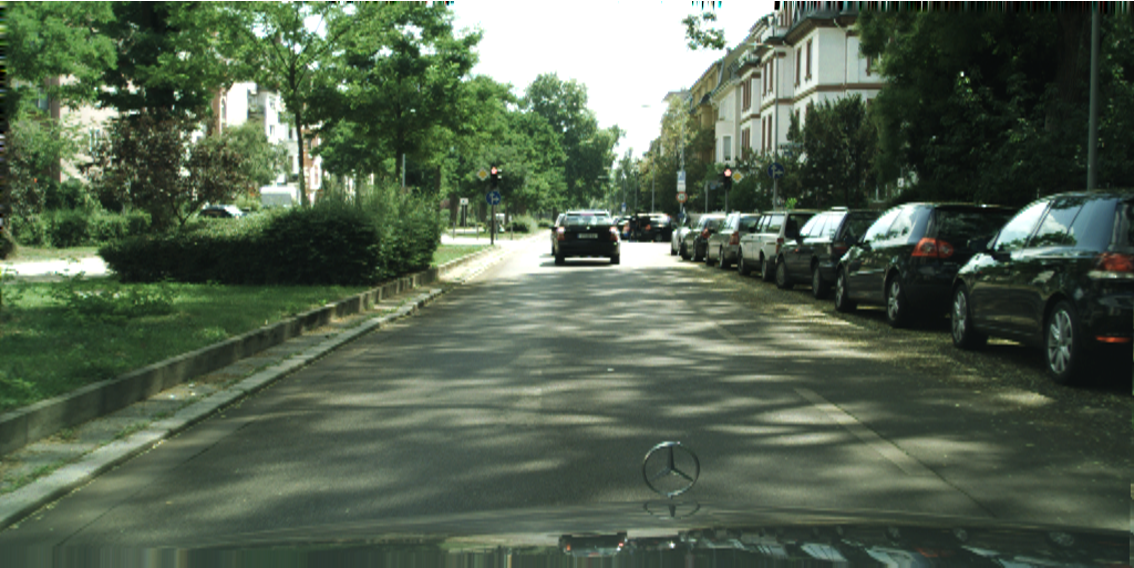} &
    \includegraphics[width=0.45\columnwidth,height=0.06\textheight]{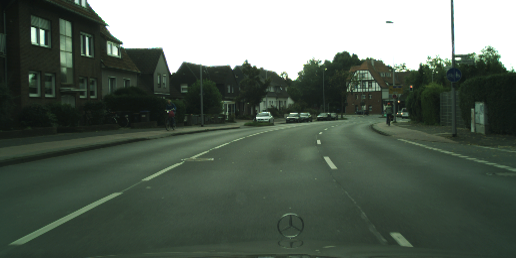} \\
    \multicolumn{2}{c}{Reference images}\\
    \includegraphics[width=0.45\columnwidth,height=0.06\textheight]{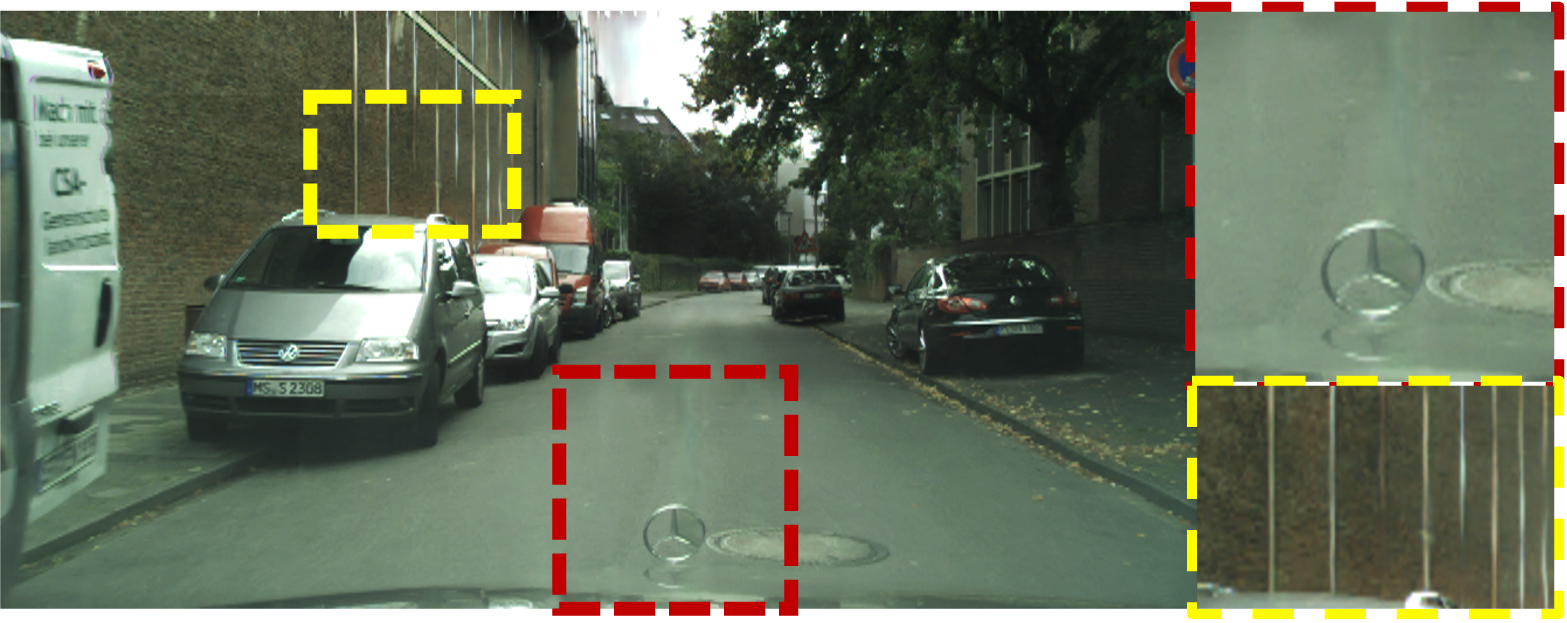} &
   \includegraphics[width=0.45\columnwidth,height=0.06\textheight]{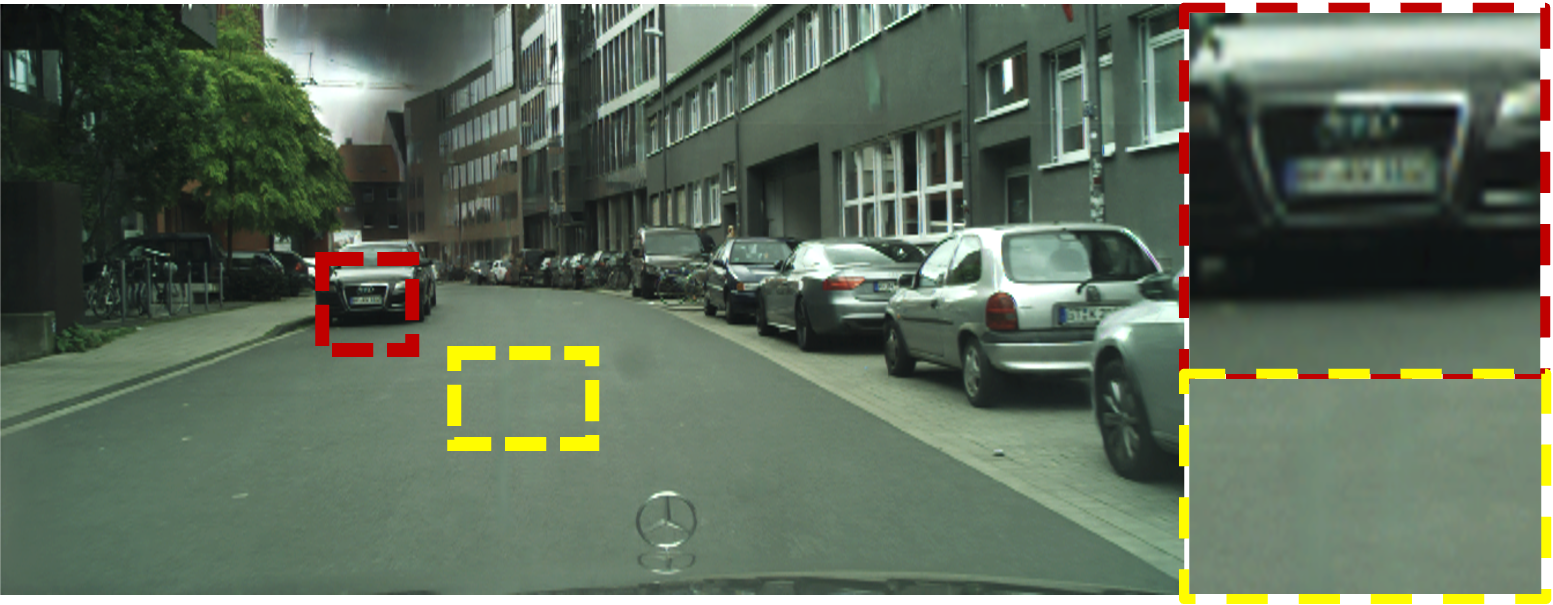}\\
    \multicolumn{2}{c}{Ours} \\
   \includegraphics[width=0.45\columnwidth,height=0.06\textheight]{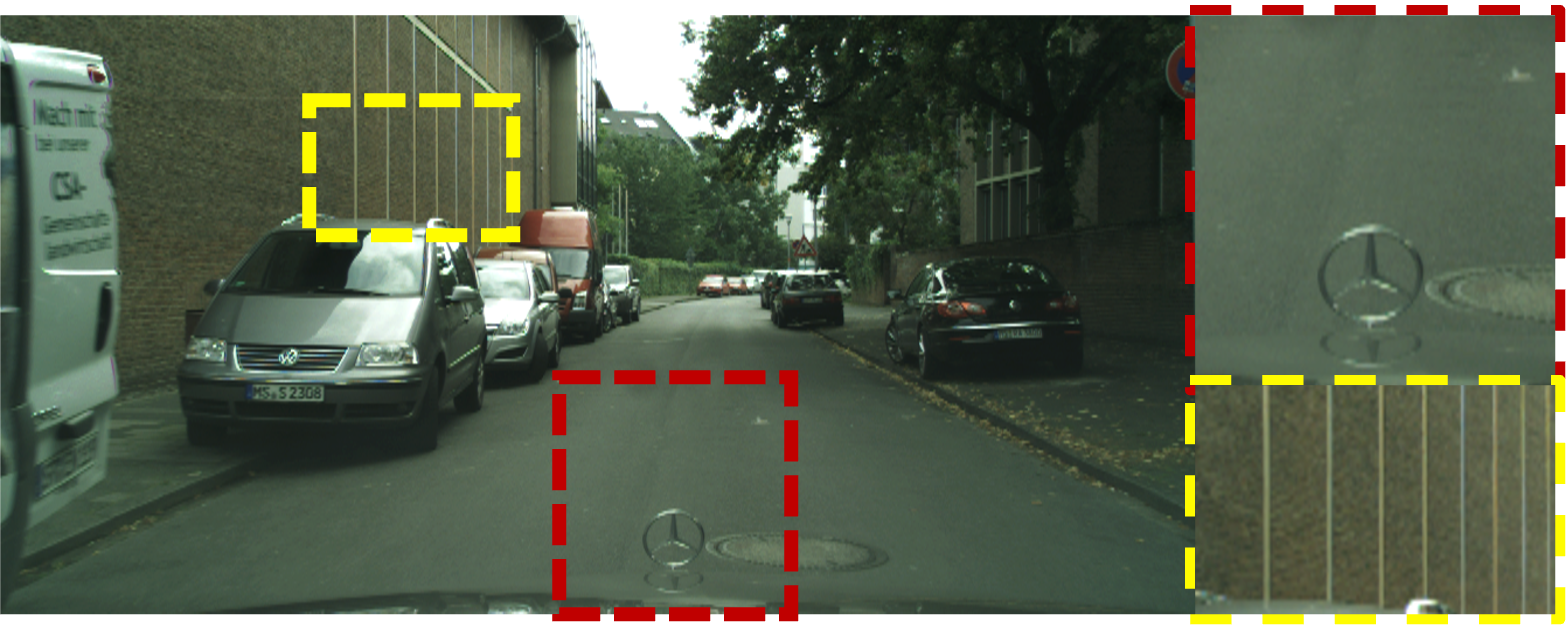} &
   \includegraphics[width=0.45\columnwidth,height=0.06\textheight]{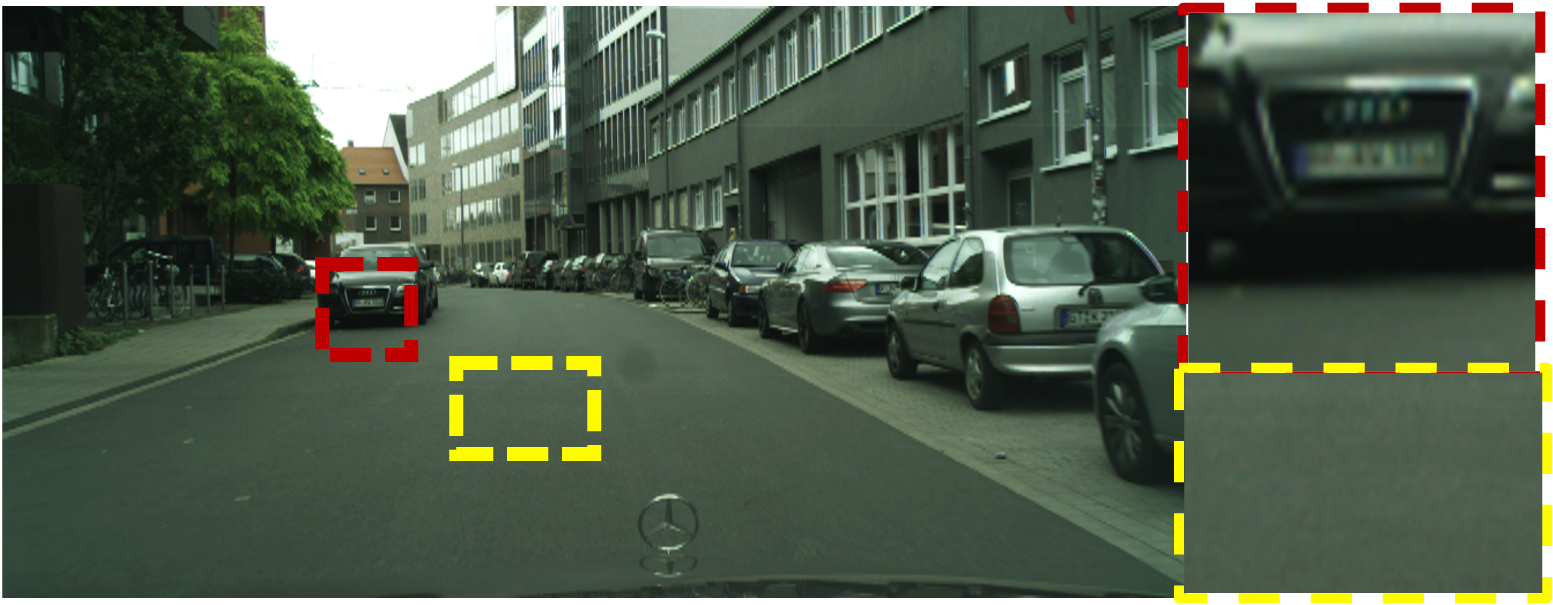}\\
   \multicolumn{2}{c}{Ground truth} \vspace{-9pt}
  \end{tabular}
  \caption{Sample de-rained images from Cityscapes-Rain~\cite{tremblay2020rain}. Unlike existing methods, our reference-guided de-raining filter enhances the de-rained results using a reference clean image as guidance. \vspace{-9pt} \vspace{-5pt}
  % Reference-guided image de-raining. Given a baseline model that performs de-raining on a rainy image, we propose to use a reference image as guidance to further enhance the de-raining result. 
}
  \label{fig:1}
\end{figure}

\begin{figure*}
\centering
\includegraphics[width=1\linewidth]{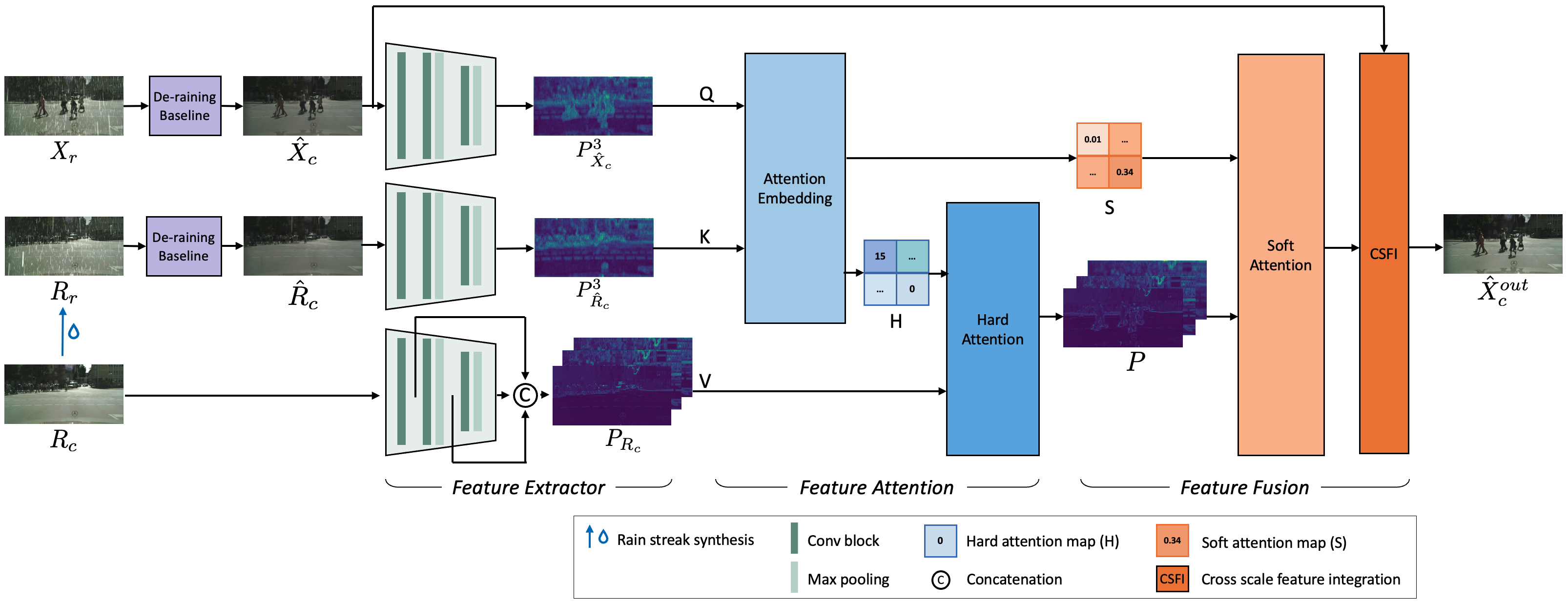}
\caption{
Overview of our framework. We first obtain a input rainy image, $X_r$, and a synthesized reference rainy image, $R_r$. Using an existing de-raining model, we obtain the input de-rained image, $\hat X_c$, and the reference de-rained image, $\hat R_c$. These two de-rained images and the reference clean image, $R_c$ are used as input to our reference-guided de-raining filter. By capturing the useful information from the features from $R_c$, and transferring it to $\hat X_c$, we can generate the enhanced de-raining output, $\hat X_c ^{out}$.
}\label{fig:2}
\end{figure*}

% Firstly, they are projected into feature space in the Feature Extractor module, resulting in $P^3_{\hat X_c}$, $P^3_{\hat R_c}$ and $P^3_{ X_c}$. Feature Attention module will further calculate relevance between query $P^3_{\hat X_c}$ and key $P^3_{\hat R_c}$. And output useful feature $P$ taking $P_{R_c}$ as the value according to the hard attention map $H$. $P$ and soft attention map $S$ are compensated to de-rained image $\hat X _c$ to get the pipeline output $\hat X_c ^{out}$

In this paper, we propose a novel framework for image de-raining. We use existing de-raining models as baselines and present a reference-guided de-raining filter that extracts useful feature information from a reference clean image to compensate for the baseline results. The key insight is to transfer useful features from a reference clean image. Our framework consists of a feature extractor, a feature attention module, and a feature fusion module. Given a rainy image and a reference rainy image as input, we first estimate the de-rained images using an existing de-raining model. We use these results with a reference clean image as input for the feature extractor that extracts the multi-scale features of each image. The feature attention module uses these features as input to estimate useful features from a reference rain/clean image. At this point, the feature attention module computes the most relevant feature patch from the reference clean image. Finally, we introduce the feature fusion module to aggregate multi-scale features.

We summarize our main contributions as follows:

\begin{itemize}[leftmargin=2em]
\setlength{\parskip}{1pt}
\item We propose a novel framework to integrate existing de-raining methods into a reference-guided de-raining filter that captures useful features by leveraging reference images.
\item Our method can be used with a wide range of existing methods in a plug-and-play manner.
\item Experimental results show that our method improves the performance of existing methods, from a prior-based to a state-of-the-art method.

\end{itemize}

\section{Problem Formulation}
\label{sec:problem}
% \CO{XX notations can be improved: e.g. we have target rainy image (), target derained image (), reference rainy image (), reference derained image () and final enhancement output (). The notations should be relevant according to their relationshipXX}
% Existing data-driven single image de-raining models aim to construct frameworks to perform a de-raining transform $f_{\theta_S}(\cdot)$ parameterized by ${\theta_S}$. We denote original rainy images that need to be de-rained as $X$, and the corresponding target clean images as $Y$. In this way, de-rained images can be denoted as $\hat X \triangleq f_{\theta_S}(X)$. The goal of single image de-raining models is to optimize the transform so that $\hat X$ can approach $Y$ given the distance function $\mathcal{L}$, which is expressed as: 
% %
% \begin{equation}
% \mathop{\arg\min}\limits_\theta \ \mathcal{L}(f_{\theta_S}(X), Y)
% \end{equation}
% %

% In this paper, we also consider reference clean images $R$ containing similar scenes to $X$. Our model represented as $g_{\theta_R}$ which parameterized by ${\theta_R}$ aims to extract useful pattern information from $R$ and compensate for the original de-rained image $\hat X$, resulting final compensated de-rained image $g_{\theta_R}(\hat X, R)$. Given the loss function $\mathcal{L}$, our optimization can be formulated as 
% %
% \begin{equation}
% \mathop{\arg \min}\limits_{\theta_R} \ \mathcal{L}(g_{\theta_R}(\hat X, R), Y) = \mathop{\arg \min}\limits_\Theta \ \mathcal{L}(g_{\theta_R}(f_{\theta_S}(X), R), Y)
% \end{equation}%

Given an input rainy image $X_r$, existing learning-based single image de-raining methods aim to learn a model $f_{\theta_S}(\cdot)$, parameterized by ${\theta_S}$, that can generate an estimated clean image $\hat X_c = f_{\theta_S}(X_r)$ in which the rain streaks are removed. The model $f_{\theta_S}(X_r)$ is learned by minimizing the error between $\hat X_c$ and the ground-truth clean image $X^{gt}_c$ using a loss function $\mathcal{L}$ as:

% Existing data-driven single image de-raining models aim to perform de-raining with a model, $f_{\theta_S}(\cdot)$, parameterized by ${\theta_S}$. We denote an original rainy image $X_r$, and the corresponding clean image as $X_c$. The goal of single image de-raining models is to generate the estimation of a clean image, $\hat X_c \triangleq f_{\theta_S}(X)$, by learning parameters $\theta_S$ that minimizes a loss function, $\mathcal{L}$, as: 
%
\begin{equation}
\mathop{\arg\min}\limits_{\theta_S} \ \mathcal{L}(f_{\theta_S}(X_r), X^{gt}_c).
\end{equation}

In this paper, we propose to further employ a reference clean image $R_c$ as guidance to improve the result of existing de-raining models and generate the enhancement output $\hat X^{out}_c$. We present a model $g_{\theta_R}(\cdot)$, parameterized by ${\theta_R}$, that aims to extract useful feature information from $R_c$ to compensate for $\hat X_c$, resulting in the final enhancement output $\hat X^{out}_c = g_{\theta_R}(\hat X_c, R_c)$. Namely, our model learns to minimize the following loss function:  
\begin{equation}
\mathop{\arg \min}\limits_{\theta_R} \ \mathcal{L}(g_{\theta_R}(\hat X_c, R_c), X^{gt }_c).
\end{equation}

\section{Method}
\label{sec:method}
Figure~\ref{fig:2} shows the overview of our framework. The proposed reference-guided de-raining filter (RDF), $g_{\theta_R}(\cdot)$, is designed to extract useful feature information from $R_c$ to compensate for $\hat X_c$ obtained from an existing baseline model. 
Given $\hat X_c$, we first perform image retrieval to find $R_c$ from an image database. We then obtain the synthesized rainy image $R_r$ by synthesizing the rain streaks to $R_c$~\cite{CHOI2022421}. By using $R_r$ as input to the baseline model, we can estimate the reference de-rained image $\hat R_c$. By capturing the similarity between the two de-rained images, $\hat X_c$ and $\hat R_c$, RDF aims to transfer the useful information from a reference clean image $R_c$ to $\hat X_c$, generating the enhancement output $\hat X^{out}_c$.

RDF mainly consists of three components: feature extractor, feature attention and feature fusion.  
The feature extractor first projects the images $\hat X_c$, $\hat R_c$, and $R_c$ into the features $P_{\hat X_c}$, $P_{\hat R_c}$, and $P_{R_c}$, respectively. The feature attention module firstly computes the attention weights taking $P_{\hat X_c}$ as query (Q) and $P_{\hat R_c}$ as key (K).  Attention embedding module outputs the highest relevance as soft attention maps $S$ at patch level and indexes corresponding to highest relevance at patch level as hard attention maps $H$. In the hard attention module, $H$ is further used to select the most relevant patch from the paired value (V), $P_{R_c}$, estimating the feature $P$, the useful feature extracted from the reference clean image. In the feature fusion module, $P$ is re-weighted at the patch level using the soft attention maps $S$. The re-weighted feature is then integrated with the de-rained image through the Cross-Scale Feature Integration (CSFI) stage. Finally, the fused feature is back-projected into the image space to produce the enhancement output $\hat X^{out}_c$.

\begin{figure}[t]
  \centering
  \setlength{\tabcolsep}{2pt} 
  \begin{tabular}{ccc}
    \includegraphics[width=0.32\columnwidth]{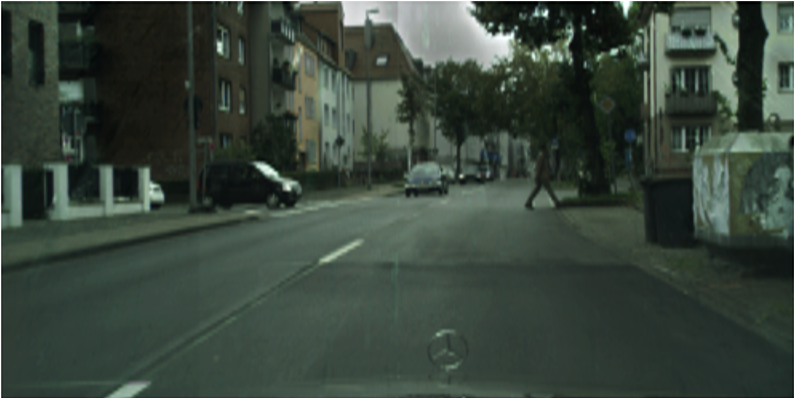} &
    \includegraphics[width=0.32\columnwidth]{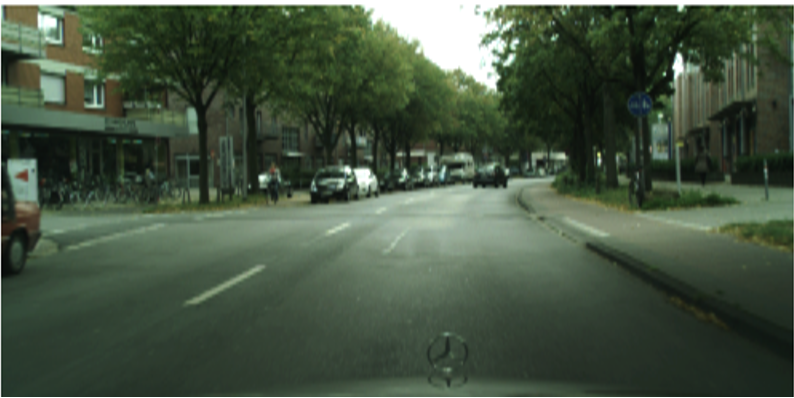} &
    \includegraphics[width=0.32\columnwidth]{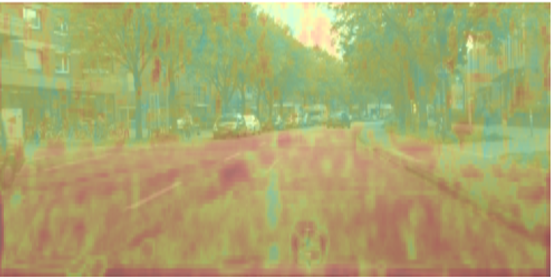} \\
    
    \includegraphics[width=0.32\columnwidth]{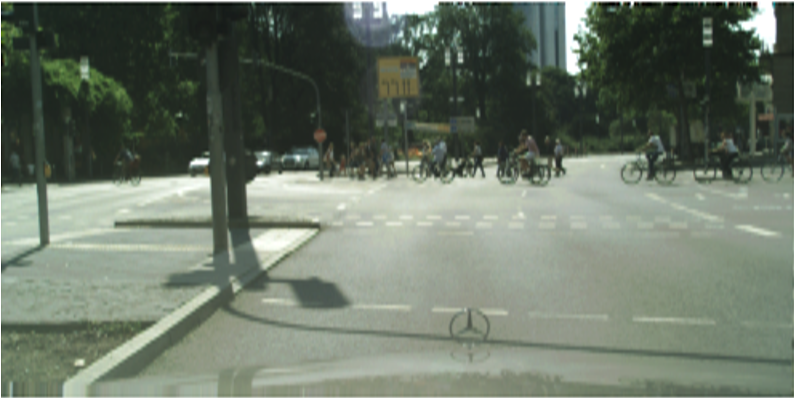} &
    \includegraphics[width=0.32\columnwidth]{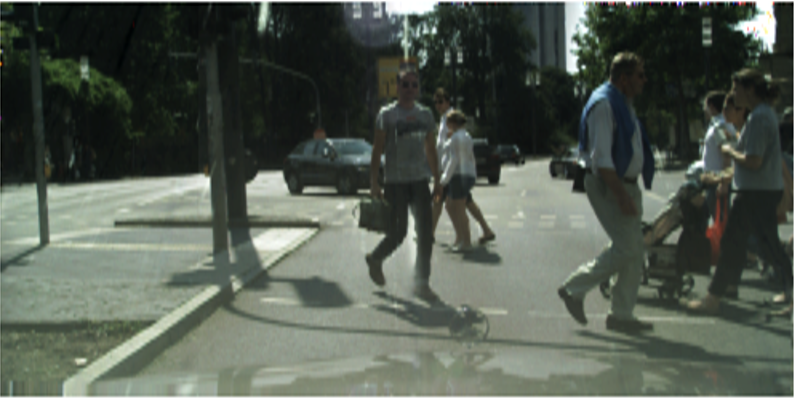} &
    \includegraphics[width=0.32\columnwidth]{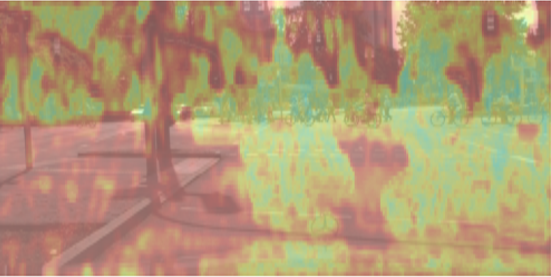} \\
    
    \includegraphics[width=0.32\columnwidth]{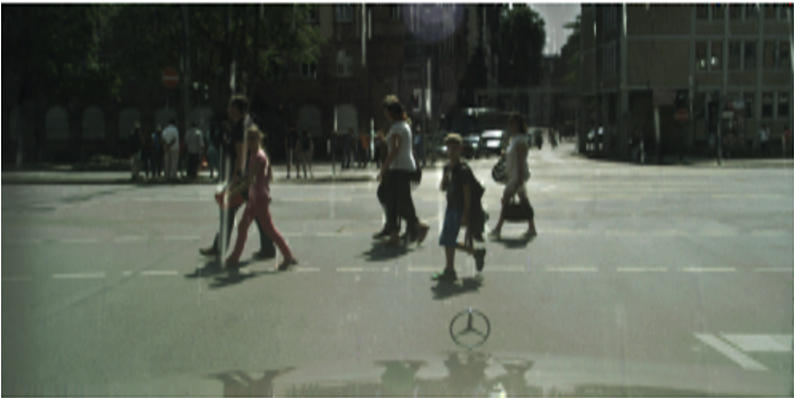} &
    \includegraphics[width=0.32\columnwidth]{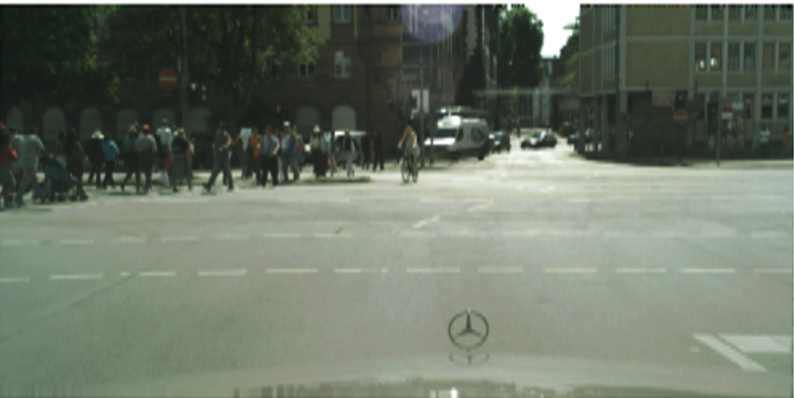} &
    \includegraphics[width=0.32\columnwidth]{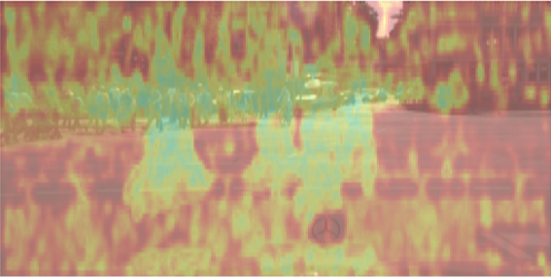} \\
    % (a) $\hat{X}_c$ & (b) $\hat{R}_c$ & (c) Attention map 
    $\hat{X}_c$ & $\hat{R}_c$ & Attention map 
  \end{tabular}
  \caption{Attention maps from the feature attention module. Using the feature of a input de-rained image, $\hat{X}_c$, as a query, and the feature of a reference de-rained image, $\hat{R}_c$, as a key, we compute attention weights that are utilized to select the useful features from the reference image. 
  The attention maps are color-coded, where warmer colors indicate higher values.
  % Attention maps are color-coded from blue (smaller), yellow to red (larger).
}
  \label{fig:attn map}
\end{figure}

\subsection{Feature extractor}
% \begin{figure}[!t]
%     \centering
%     \includegraphics[width=0.8\linewidth]{}
%     \caption{pattern Extractor module: At this stage, images are projected into three levels of patterns. Specifically, Level 1, Level 2, and Level 3 patterns are represented in the shapes of $(B,C,H,W)$, $(B,2C, H/2, W/2)$, and $(B, 4C, H/4, W/4)$, respectively.
% }
%     \label{}
% \end{figure}
The feature extractor module, including several convolution blocks, is designed to project images into a feature space. Specifically, the feature extractor maps a single image into three distinct feature levels. The Level-1 feature,  $P_{\{\cdot\}}^1 \in \mathbb{R}^{(B, C, H, W)}$, preserves the input image size but with a higher-dimensional channel space $C$. The Level-2 and Level-3 features are represented at lower resolutions and increased channel dimensions, specifically denoted as $P_{\{\cdot\}}^2 \in \mathbb{R}^{(B, 2C, H/2, W/2)}$ and $P_{\{\cdot\}}^3 \in \mathbb{R}^{(B, 4C, H/4, W/4)} $, respectively. In this context, the feature $ P_{\{\cdot\}} \triangleq \{ P_{\{\cdot\}}^1, P_{\{\cdot\}}^2, P_{\{\cdot\}}^3\} $ aggregates these three-level features extracted from an input of the feature extractor.

\subsection{Feature attention}
The feature attention module receives three different features, \( P_{{\hat X}_c} \), \( P_{R_c} \), and \( P_{{\hat R}_c} \). This module maps the query \( P_{{\hat X}_c} \) to useful feature patches in the context of the key \( P_{{\hat R}_c} \) and value \( P_{ R_c} \), subsequently outputting the useful features \( P \) along with the corresponding relevance maps at the patch level. During the attention embedding stage, only the Level-3 features \( P_{{\hat X}_c}^3 \) and \( P_{{\hat R}_c}^3 \) are utilized to compute relevance for patches, as they encapsulate more abstract information and a wider receptive field. Following this, the hard attention module selects the most relevant feature patch of \( P_{R_c} \) for the patches of \( P_{{\hat X}_c} \), using the relevance just computed at every feature level.
In Figure~\ref{fig:attn map}, we visualize the attention maps, where we could see that similar areas are highly noticed, which are used to compensate for de-rain results later.

\subsection{Feature fusion}
\begin{figure}[t]
    \centering
    \includegraphics[width=1\linewidth]{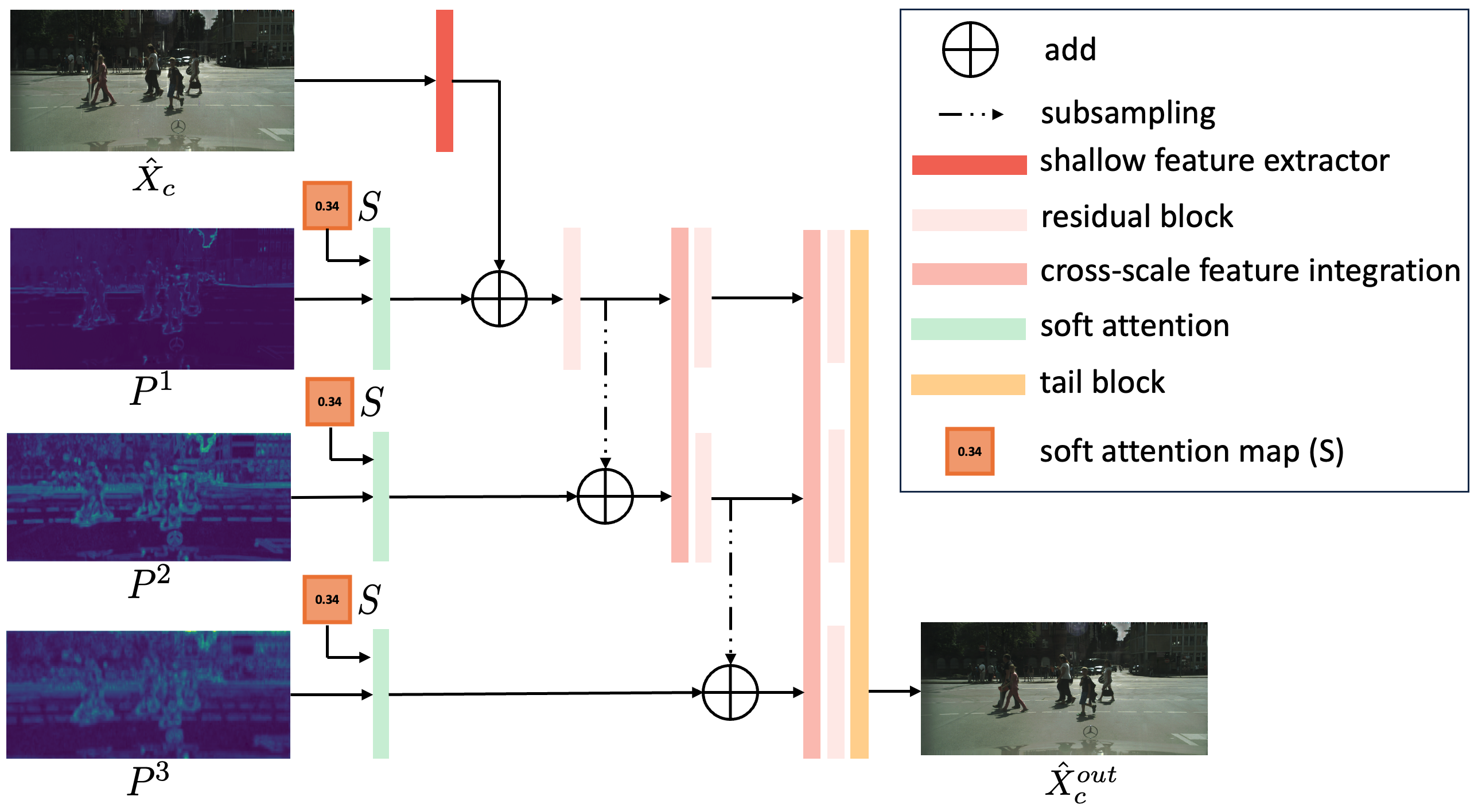}
    \caption{Feature fusion module. The de-rained images are first projected into the feature space using a shallow feature extractor. The features at each level are then compensated sequentially from level 1 (fine-level) to Level 3 (coarse-level). }
    \label{fig:PF}
\end{figure}
In Figure~\ref{fig:PF}. the CSFI module represents a well-established method for blending features with various scales~\cite{yang2020learning,sun2019highresolution}. In this paper, we leverage this approach to fuse $P$, the useful features extracted from the reference, with the original de-rained image. The useful feature is incrementally added to the original de-rained image $ \hat X $, guided by relevance maps. During the compensation stage, the CSFI module is employed in conjunction with the residual block to facilitate information sharing across all feature levels. The level-1 feature, characterized by its relatively precise information and intricate details, is compensated to the image first. Conversely, the level-3 feature, which encapsulates more abstract information, is compensated to the image last. This systematic process results in a projection from the feature space to the image space, generating the final output $\hat{X}_{c}^{out}$.

% \begin{figure}[!t]
%     \centering
%     \includegraphics[width=1\linewidth]{TF.png}
%     \caption{pattern Fusion Module: At this stage, de-rained images are first projected into the pattern space using a shallow feature extractor. patterns at each level are then compensated in a sequence from the more precise Level 1 to the more abstract Level 3. This compensation is performed after the soft attention module, utilizing the soft attention maps \( S \) obtained from the pattern Attention stage.
% }
%     \label{fig:enter-label}
% \end{figure}

\begin{table*}[!t]
\centering
\caption{Quantitative evaluation on the three datasets: BDD100K-Rain, synthesized using SyRaGAN~\cite{CHOI2022421} and BDD100K~\cite{bdd100k},  Cityscapes-Rain~\cite{tremblay2020rain}, and KITTI-Rain~\cite{tremblay2020rain}. Experiments include a prior-based model, GMM~\cite{li2016rain}, a CNN-based model, PReNet~\cite{ren2019progressive}, and a transformer-based model, Uformer~\cite{Wang_2022_CVPR}, and their improvement using our module highlighted in blue.}

\resizebox{0.9\linewidth}{!}{
\setlength{\tabcolsep}{3mm}
\begin{tabular}{ r ll ll ll}

\specialrule{1.2pt}{0.2pt}{1pt}
\multicolumn{1}{c}{Methods} & \multicolumn{2}{c}{BDD100K-Rain} & \multicolumn{2}{c}{KITTI-Rain} & \multicolumn{2}{c}{Cityscapes-Rain}  \\
\cmidrule(lr){1-1}
\cmidrule(lr){2-3}
\cmidrule(lr){4-5}
\cmidrule(lr){6-7}

Name        & PSNR  &SSIM     & PSNR &SSIM       & PSNR &SSIM  \\ 

\cmidrule(lr){1-7}
GMM~\cite{li2016rain}   & $28.37$ & $0.8590$  & $17.08$ & $0.4818$       & $23.333$ &  $0.7830$ \\
PReNet~\cite{ren2019progressive}    & $33.38$ & $0.9474$  & $22.71$ & $0.7497$   & $23.80$ & $0.9529$ \\
Uformer~\cite{Wang_2022_CVPR}    & $36.30$ & $0.9619$   & $31.59$ & $0.9694$           & $23.98$ & $0.9509$ \\ 
\cmidrule(lr){1-7}
GMM + \textbf{Ours}   & $31.44_{\note{+3.07}}$ & $0.9003_{\note{+0.0412}}$ & $25.23_{\note{+8.15}}$ & $0.7933_{\note{+0.3114}}$     & $23.48_{\note{+0.14}}$ & $0.8869_{\note{+0.1039}}$ \\
PReNet + \textbf{Ours} & $33.72_{\note{+0.34}}$ & $0.9487_{\note{+0.0013}}$  & $26.92_{\note{+4.21}}$ & $0.8551_{\note{+0.1054}}$     & $24.98_{\note{+1.18}}$ & $0.9595_{\note{+0.0066}}$ \\
Uformer + \textbf{Ours} & $36.37_{\note{+0.07}}$ & $0.9627_{\note{+0.0008}}$    & $33.05_{{\note{+1.46}}}$ & $0.9761_{{\note{+0.0067}}}$     & $25.64_{\note{+1.66}}$ & $0.9601_{\note{+0.0091}}$ \\
\specialrule{1.2pt}{0.2pt}{1pt}
\end{tabular}
}
\label{tab:overall}
\end{table*}

\subsection{\textbf{Loss}}
During the training phase, our approach consists of initialization and fine-tuning stages. In the initialization stage, the model is trained to transfer useful features from clean object images using an L1 loss function. The model learns the transformation from a derained to a clean image in this stage, ensuring the model has the capability to perform deraining on simple rainy images. In the fine-tuning stage, the model is further trained to transfer useful features from reference clean images to better simulate real-world application scenarios. For the independent objectives of these two stages, we employ the loss functions tailored to each stage's specific requirements. During the initialization stage, the loss function is defined as the $\ell1$ reconstruction loss:
\begin{equation}
\mathcal{L} (\hat{X}_{c}^{out}, X_c) = \| \hat{X}_{c}^{out} - {X_{c}} \|_1.
\end{equation}
In the fine-tuning stage, we apply the MS-SSIM-L1-Loss~\cite{7797130}:
\begin{equation}
\mathcal L (\hat{X}_{c}^{out}, X_c) = \alpha_1 \| \hat{X}_{c}^{out} - X_{c}\|_1 + \alpha_2 (1 - \text{SSIM}(\hat{X}_{c}^{out}, X_c)), 
\end{equation}
where \(\alpha_1\) and \(\alpha_2\) are the hyperparameters to control the effect of each loss.

\section{Validation}
\label{sec:validation}

\subsection{Experiment setup}
\label{ssec:setup}
% \textbf{Baseline. }
In our framework, we use existing de-raining models as baselines. We adopted three baseline models, including the prior-based (GMM~\cite{li2016rain}), CNN-based (PReNet~\cite{ren2019progressive}), and transformer-based (Uformer~\cite{Wang_2022_CVPR}) models. Except for the prior-based method that does not require training, we used public codes for training each baseline on each dataset.

For the dataset, we require both a clean/rainy image pair and a reference clean/rainy image pair that contain similar scenes. However, existing rain benchmarks, such as Rain100L~\cite{9157472} and DID~\cite{8578177}, have limited similar scenes, providing unreasonable reference images. Therefore, we constructed the dataset as follows:
\begin{itemize}[leftmargin=2em]
\setlength{\parskip}{1pt}
\item BDD100K-Rain: We used BDD100K~\cite{bdd100k}, a large-scale driving scene dataset, and synthesized the rain streaks by using SyRaGAN~\cite{CHOI2022421} and obtained $256\times 256$ images. 
\item Cityscapes-Rain/KITTI-Rain: This dataset includes $256\times 1024$ images and constructed by~\cite{tremblay2020rain}, which renders rain streaks to evaluate bad weather. 

\end{itemize}

For image retrieval, we implemented the reference image retrieval method at the image hash level. First, all images in the dataset are projected into the image perceptual hash space. Then, for every image that requires compensation, the nearest neighbor image is selected as the reference image.

% \JHcomment{Hardware spec with pytorch would be described and  alpha value also be described in the loss}
% \Zihaocomment{modified}
We utilized 8 A100 GPUs and PyTorch for our experiments. The channel numbers for levels 1, 2, and 3 are set to $64$, $128$, and $256$, respectively. The convolution blocks in the feature extractor module are initialized with VGG-19~\cite{simonyan2015deep}. At the feature attention stage, the patch sizes for levels 1, 2, and 3 are set to $12$, $6$, and $3$, respectively. Within the feature fusion module, the features for levels 1, 2, and 3 are configured as $(64, H, W)$, $(128, H/2, W/2)$, and $(256, H/4, W/4)$, respectively. To ensure that our model can capture useful features and compensate accurately, we initially employ the input clean image $X_c$ as the reference image for initialization training using the loss functions in (3). The initialization training is done for each model on each dataset. After the initialization training stage, $X_c$ is replaced with the actual reference image in the subsequent training stage for fine-tuning. In the fine-tuning stage, we set $\alpha_1 = 0.6$ and $\alpha_2 = 0.4$ and use the loss function in (4).

\subsection{Discussion}
\label{ssec:comparison}
Table~\ref{tab:overall} presents the quantitative results of the baselines and the improvements achieved using our method. Although each baseline employs a different methodology for image de-raining, our method can universally enhance the performance of these baseline models. This suggests that our reference-guided de-raining filter effectively extracts useful features from reference images. The improvement is most significant on the KITTI-Rain dataset, as this dataset provides better reference images. Our pipeline also demonstrates various degrees of compensation effects on different baselines. For Uformer \cite{Wang_2022_CVPR}, which achieves the best results among the three baseline models, our method shows relatively small improvements. On the other hand, for GMM~\cite{li2016rain}, which is the earliest method among the three baseline models, our method shows more substantial improvements.

% In this paper, we propose a Reference-guided de-raining filter designed to compensate for useful features from similar scenes. We highlight that many de-raining application scenarios occur in environments containing similar scenes, which can be leveraged to enhance the performance of existing de-raining models. Through extensive experiments on various baselines, each based on different architectures or priors, and across different datasets, we demonstrate the effectiveness of our approach. Comparison results show that our model is capable of improving the de-raining results of baseline models in terms of both \( PSNR \) and \( SSIM \). An ablation study of the reference images further substantiates that our model operates by transferring useful features. This work opens new avenues for exploring feature-based compensation techniques in image de-raining and potentially other image restoration tasks.

% \subsection{Analysis}
% \label{ssec:analysis}

		\begin{table}[!t]
		\centering
		\caption{Effect of reference images. PReNet~\cite{ren2019progressive} trained on BDD100K-Rain~\cite{bdd100k, CHOI2022421} is used as a backbone while changing the reference images to the ground truth clean image, Gaussian noise image, and our reference image obtained by image retrieval. 	}
			\vspace{5pt}
\scriptsize
\resizebox{0.95\linewidth}{!}{
\setlength{\tabcolsep}{12pt}
			\begin{tabular}[b]{c|cc}
			\toprule
		 Reference Type& 	PSNR  & SSIM      \\   			\hline
			 Ground truth   &   $35.50_{\note{+ 2.06}}$ &    $ 0.9736_{ \note{+ 0.0257}}$ \\
			Noise image  & $ 33.37_{\noter{- 0.07}}$ & $0.9470_{\noter{- 0.0009}} $\\
   Reference image   &   $33.78_{\note{+ 0.34}}$  &  $0.9491 _{\note{+ 0.0013}}$   \\
			\bottomrule
		\end{tabular}}
  
			\label{tab:references}
		\end{table}
  
We further evaluate the effect of reference images on the BDD100K-Rain dataset, specifically to analyze how a reference image can contribute to the de-raining process. Using PReNet as a baseline model, we analyze the effect of reference images using three different image types: ground truth clean images, Gaussian-noise images, and our reference images collected using image retrieval. As shown in Table~\ref{tab:references}, ground truth images, which encapsulate all the useful information, yield the best results and significantly enhance the performance. Noise images, that include less relevant information, produce the worst outcomes. Reference images containing similar scenes provide results that fall between the upper bound set by the ground truth images and the lower bound established by the noise images. The results indicate that our model can transfer useful features from reference images, and the degree of enhancement primarily depends on the quality of the reference images.

\begin{figure}[t]
\centering
\begin{tabular}{c@{\hspace{2pt}}c@{\hspace{2pt}}c}  
\includegraphics[height=0.05\textheight]{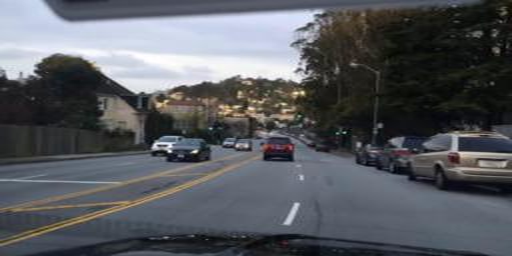}
&\includegraphics[ height=0.05\textheight]{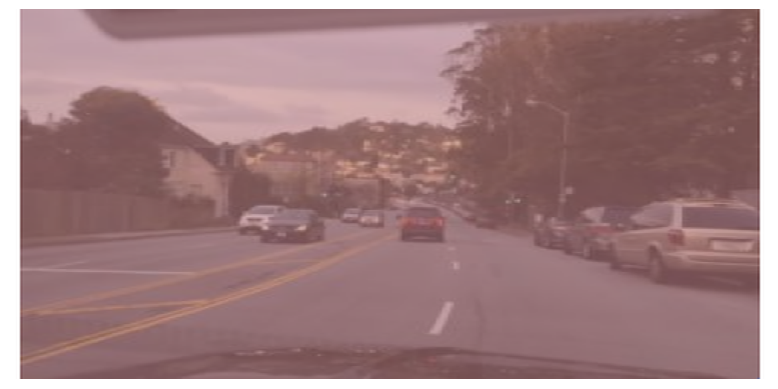}
&
\begin{overpic}[height=0.05\textheight]{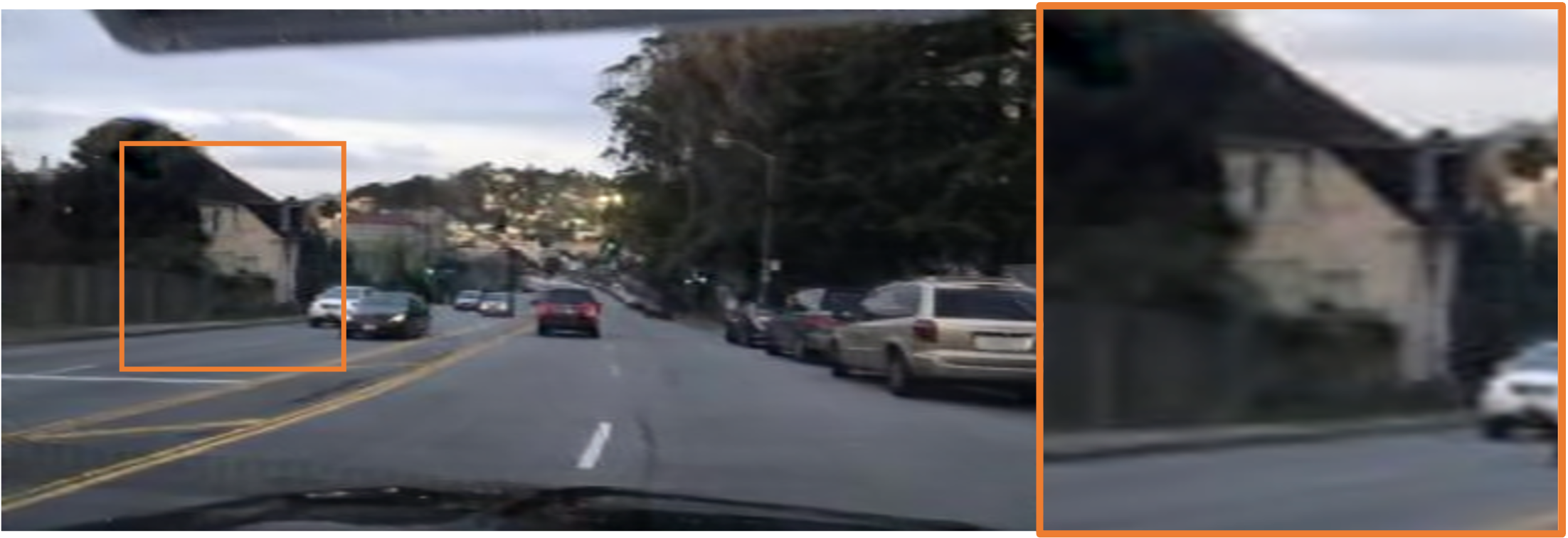} 
\put(1,1){\scriptsize \contour{black}{\protect\textcolor{white}{}}}
 \end{overpic}
\\
 
\includegraphics[ height=0.05\textheight]{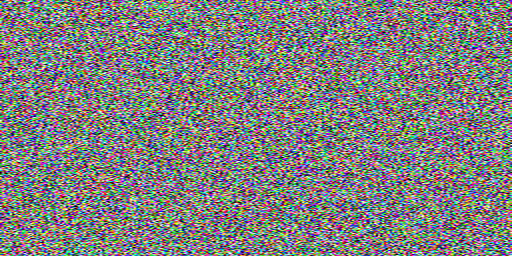}
&\includegraphics[ height=0.05\textheight]{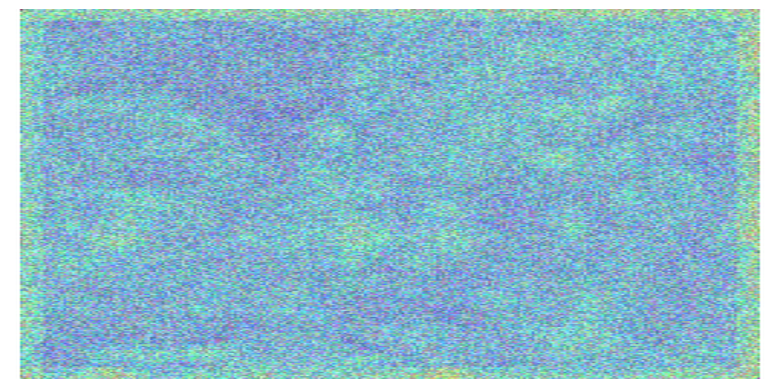}
&
\begin{overpic}[height=0.05\textheight]{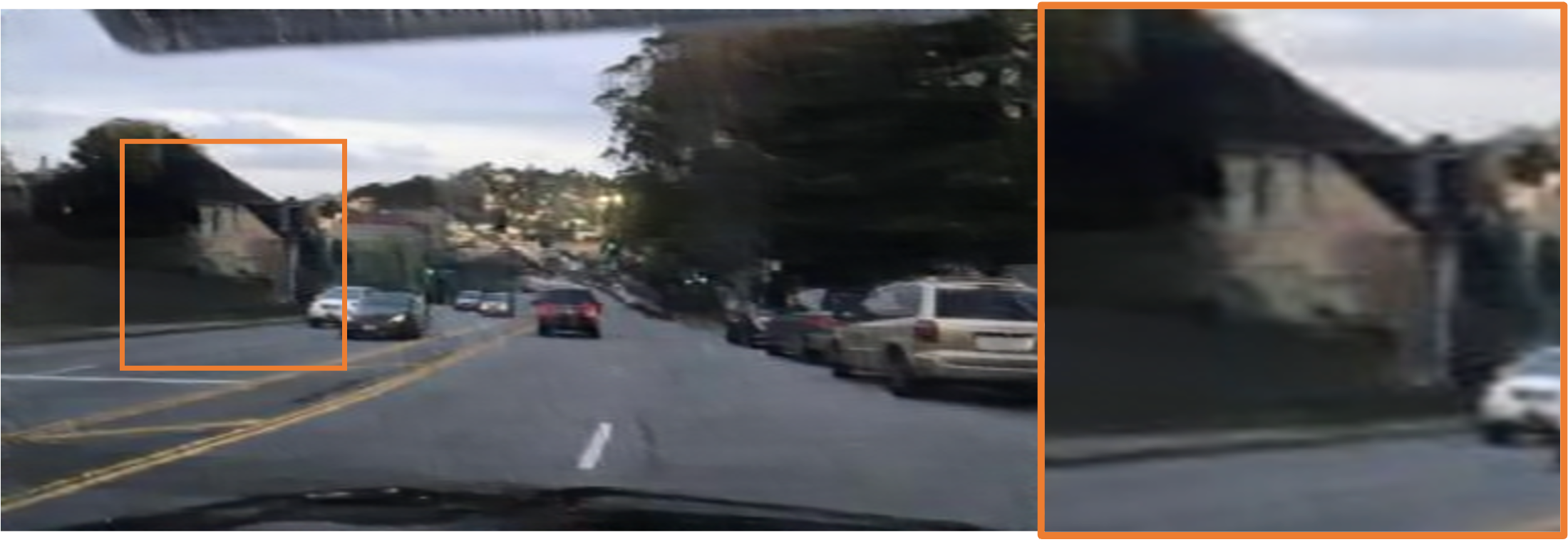} 
\put(1,1){\scriptsize \contour{black}{\protect\textcolor{white}{}}}
 \end{overpic}\\

\includegraphics[ height=0.05\textheight]{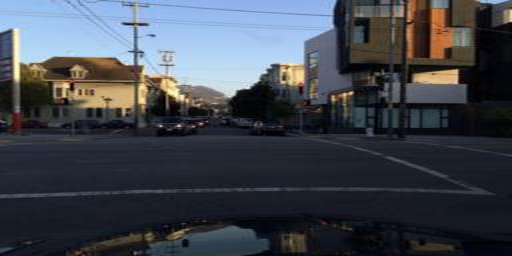}
&\includegraphics[ height=0.05\textheight]{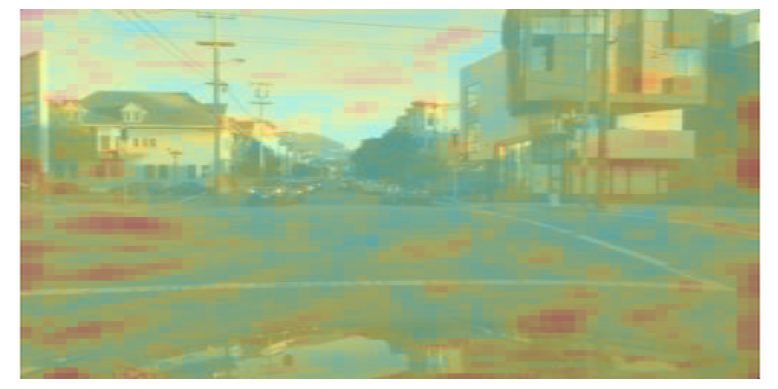}
&
\begin{overpic}[height=0.05\textheight]{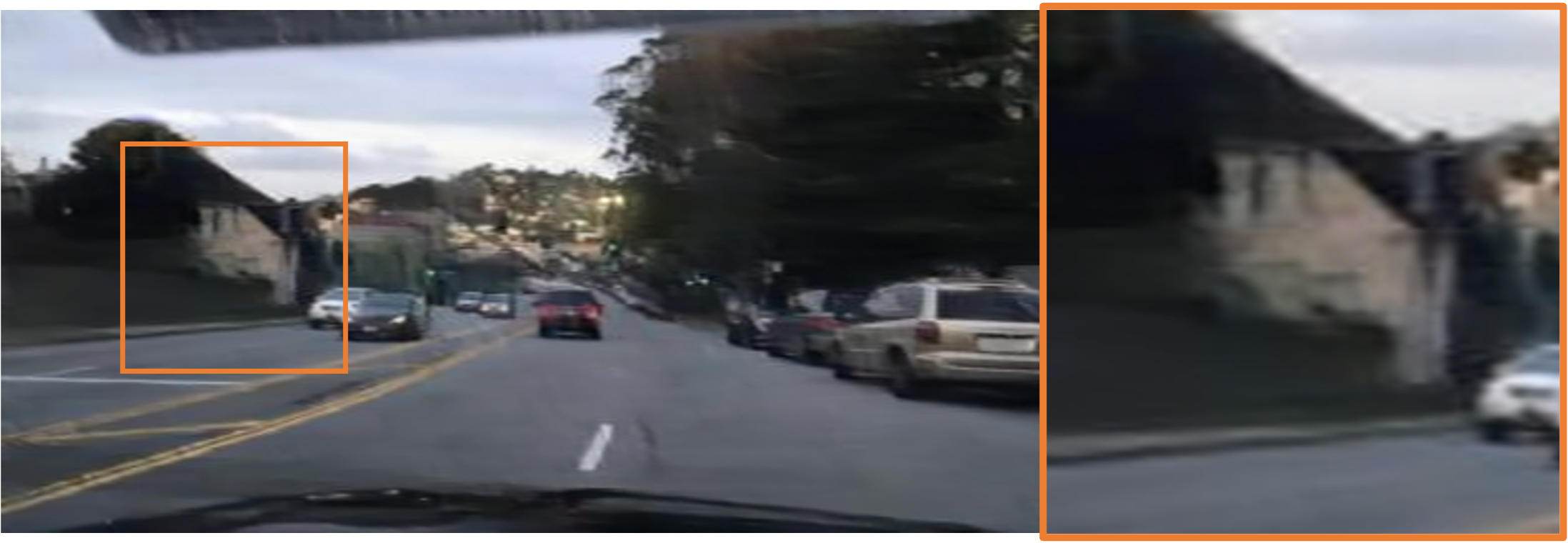} 
\put(1,1){\scriptsize \contour{black}{\protect\textcolor{white}{}}}
 \end{overpic} \\

Reference  & Attention map & De-raining output
\end{tabular}
\caption{Effect of reference images on the attention maps and de-raining results. De-raining images are obtained by using (from top to bottom) the ground-truth clean image, Gaussian noise, and our reference image obtained by image retrieval.}

\vspace{-9pt}
\label{fig:ablation}
\end{figure}

\section{Conclusion}
\label{sec:conclusion}
This paper introduces a novel framework for image de-raining that leverages a reference-guided de-raining filter, a transformer network that enhances existing de-raining results using a reference clean image as guidance. 
Our framework, as plug-and-play de-raining enhancement, shows performance improvements of prior, CNN, and transformer-based models across multiple datasets. As future work, we will integrate our method with text-to-image generation models that can synthesize clean images and use these images as references for de-raining. 

% As the task of image de-raining plays a vital role in computer vision for improving visibility and robustness in outdoor environments, our reference-guided approach offers a valuable contribution to enhancing real-world applications in this field.

% In this paper, we propose a reference-guided de-raining filter to compensate for useful features from similar scenes. 
% We point out that quite a lot de-raining application scenarios happens at the environments that containing similar scenes which could be utilize to improve de-raining performance of existing de-raining models. 
% We set extensive experiments on baselines which are based different architectures/priors on different datasets. 
% Comparison results shows that our model is capable of improving de-raining results of baseline models both on $PSNR$ and $SSIM$. Ablation study of the reference images indicates that our model works by transferring useful features. 

% TO BE added in the camera ready if the paper is accepted
% \vspace{9pt}
% \noindent
% \textbf{Acknowledgement.} This project made use of time on Tier 2 HPC facility JADE2, funded by EPSRC (EP/T022205/1).

\bibliographystyle{IEEEbib}
\bibliography{refs}

\end{document}